\documentclass{article}

\usepackage{arxiv}

\usepackage[utf8]{inputenc} 
\usepackage[T1]{fontenc}    
\usepackage{hyperref}       
\usepackage{url}            
\usepackage{booktabs}       
\usepackage{amsfonts}       
\usepackage{nicefrac}       
\usepackage{microtype}      
\usepackage{lipsum}		
\usepackage{graphicx}
\usepackage{natbib}
\usepackage{doi}
\usepackage{algorithm}
\usepackage{algorithmic}
\usepackage{amsfonts,amsmath,amssymb,amsthm}
\usepackage{booktabs}
\usepackage{multirow}
\usepackage[capitalise]{cleveref}
\usepackage{textcomp}
\usepackage{xcolor}
\usepackage{comment}
\usepackage{subcaption}
\usepackage{multicol}
\usepackage{fixltx2e}
\usepackage{relsize}
\usepackage{mathbbol}
\usepackage{xurl}
\usepackage{wasysym}
\usepackage{bm}

\title{A Comparative Study of Loss Functions: Traffic Predictions in Regular and Congestion Scenarios}

\author{Yangxinyu Xie\\
	Department of Statistics and Data Science\\
	University of Pennsylvania\\
	Philadelphia, PA \\
	\texttt{xinyux@wharton.upenn.edu} \\
	\And
	Tanwi Mallick\\
	Mathematics and Computer Science Division\\
	Argonne National Laboratory\\
	Lemont, IL \\
	\texttt{tmallick@anl.gov} \\
}

\hypersetup{
pdftitle={A Comparative Study of Loss Functions: Traffic Predictions in Regular and Congestion Scenarios},
pdfsubject={Deep Learning, Traffic Prediction, Loss Function},
pdfauthor={Yangxinyu Xie, Tanwi Mallick},
pdfkeywords={Deep Learning, Spatiotemporal Graph Neural Networks, Traffic Prediction, Loss Function}
}

\begin{document}
\maketitle

\begin{abstract}
	Spatiotemporal graph neural networks have achieved state-of-the-art performance in traffic forecasting. However, they often struggle to forecast congestion accurately due to the limitations of traditional loss functions. While accurate forecasting of regular traffic conditions is crucial, a reliable AI system must also accurately forecast congestion scenarios to maintain safe and efficient transportation.
In this paper, we explore various loss functions inspired by heavy tail analysis and imbalanced classification problems to address this issue. We evaluate the efficacy of these loss functions in forecasting traffic speed, with an emphasis on congestion scenarios. 
Through extensive experiments on real-world traffic datasets, we discovered that when optimizing for Mean Absolute Error (MAE), the MAE-Focal Loss function stands out as the most effective. When optimizing Mean Squared Error (MSE), Gumbel Loss proves to be the superior choice. 
These choices effectively forecast traffic congestion events without compromising the accuracy of regular traffic speed forecasts. This research enhances deep learning models' capabilities in forecasting sudden speed changes due to congestion and underscores the need for more research in this direction. By elevating the accuracy of congestion forecasting, we advocate for AI systems that are reliable, secure, and resilient in practical traffic management scenarios.
\end{abstract}


\section{Introduction}

With the advent of machine learning, spatiotemporal Graph Neural Networks (GNN) have emerged as a promising tool, delivering state-of-the-art results in short-term traffic speed forecasting \cite{li2018diffusion, wu2019graph, shao2022decoupled, lablack2023spatio}.  However, a significant challenge remains: current deep learning models trained using loss functions such as Mean Absolute Error (MAE) or Mean Squared Error (MSE) struggle to forecast rarer instances \cite{ding2019modeling, ribeiro2020imbalanced}. While these advanced GNNs excel at predicting regular traffic speeds, they often fall short in forecasting traffic congestion. 

\begin{figure}[t]

\centering
\includegraphics[width=0.465\columnwidth]{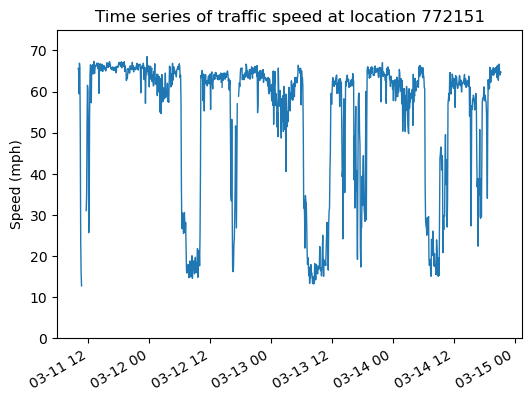}
\includegraphics[width=0.49\columnwidth]{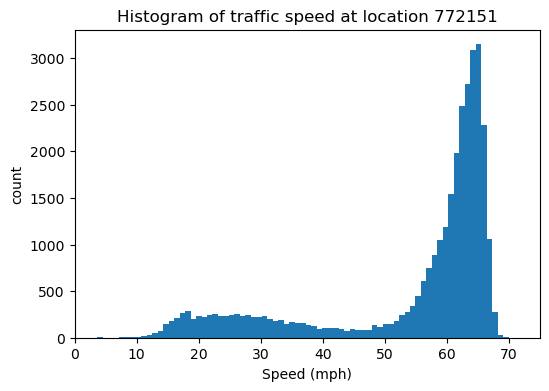}

\caption{Historical traffic speed at sensor location 772151. Left: time series from March $11^{th}$ 2012 to March $14^{th}$ 2012. Right: histogram with observations from March $1^{st}$ 2012 to June $30^{th}$ 2012.}
\label{fig: 772151}

\end{figure}

\begin{figure}[t]
\centering
\includegraphics[width=0.9\columnwidth]{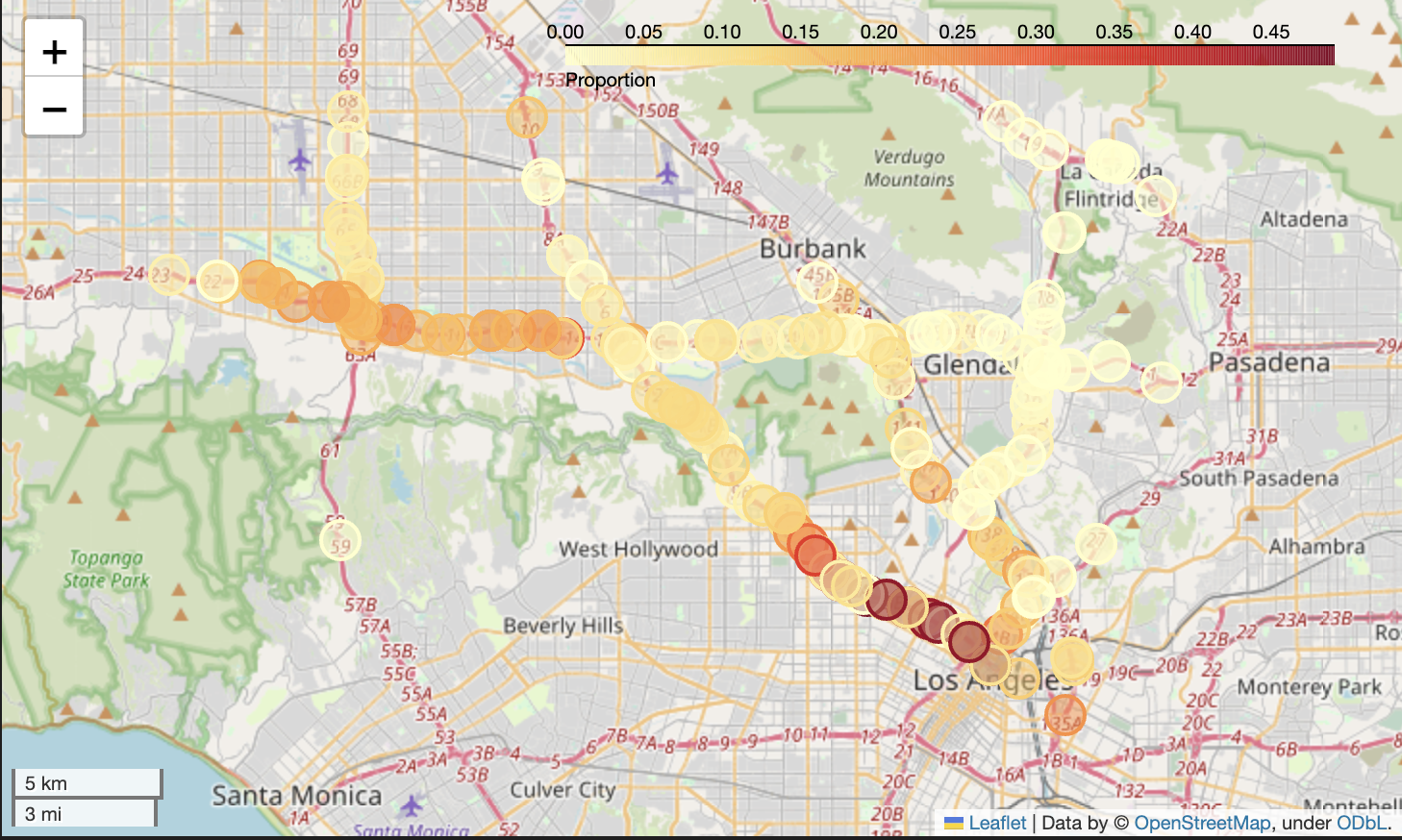}

\caption{The visualization of the levels of bimodality across different locations in the METR-LA dataset. The darker the color, the more severe the bimodality. The exact definition of the ``proportion'' in the legend is presented in the Methods section.}
\label{fig: bimodal}

\end{figure}

\cref{fig: 772151} highlights a regularly congested location in Los Angeles. Although speeds generally fluctuate between 60 and 70 mph, the traffic speed histogram reveals a pronounced bimodal distribution. This bimodality violates the normality assumptions baked into loss functions such as MAE and MSE. Moreover, as shown in \cref{fig: bimodal}, the levels of bimodality significantly differ throughout the road network, making accurate traffic speed forecasting even more complex.


Traffic congestion not only results in significant economic losses due to increased travel times and operational inefficiencies but also poses challenges in ensuring safe and efficient transportation systems. A robust and responsible AI system should accurately forecast normal, elevated, and extreme levels of congestion as it is crucial for enhancing traffic control, optimizing routing, and identifying innovative solutions to evolving patterns of congestion \cite{bishop2005intelligent,tang2005traffic,teklu2007genetic}. 

In the literature of time series forecasting, researchers have proposed innovative loss functions based on extreme value theory to better account for rare events \cite{ding2019modeling, zhang2021enhancing, kozerawski2022taming}. However, traffic congestion does not align perfectly with this theory: unlike extreme weather or pollution, traffic congestions occur more regularly throughout the year, as seen in \cref{fig: 772151}, and possess structured seasonal and spatial dependencies that often violate the assumptions of the extreme value theory. Other loss functions, inspired by imbalanced classification problems in computer vision, have been proposed \cite{ribeiro2020imbalanced, yang2021delving, ren2022balanced}. These loss functions aim to restore a balanced prediction from imbalanced training samples. However, imbalanced regression is still in an early stage and lacks an effective approach. Furthermore, these loss functions have only been tested in vision-related tasks.  

Nevertheless, these loss functions may enhance current deep learning models' abilities to predict abrupt traffic speed changes due to congestion. Unfortunately, no studies have conducted a systematic comparison of these loss functions. Moreover, most of these loss functions have not been previously employed in traffic forecasting scenarios.

In this paper, we contribute to the practice of safe, robust, and responsible AI-based systems by addressing this gap. We incorporate eight loss functions from multiple studies into two state-of-the-art spatiotemporal graph neural networks. We evaluate their efficacy in forecasting traffic speed for highway networks using three distinct metrics. These metrics will not only assess the overall performance of these loss functions but also their efficacy in pinpointing traffic congestion, ensuring a more reliable and robust AI system for real-world traffic management applications. \footnote{The source code is available at \url{https://github.com/Xieyangxinyu/A-Comparative-Study-of-Loss-Functions-Traffic-Predictions-in-Regular-and-Congestion-Scenarios}}

\section{Related Work}

\textbf{Spatiotemporal GNNs in Traffic Forecasting.}
Urban traffic inherently exhibits a non-Euclidean topological structure. To decode this structure, spatiotemporal GNNs capture spatial dependencies via a diffusion process on graphs \cite{li2018diffusion, wu2019graph} and temporal patterns using sequence-to-sequence recurrent neural networks \cite{2014GRU, 2016TCN}. Subsequent research has refined these models on benchmark datasets like METR-LA and PEMS-BAY \cite{2018GaAN, yu2018spatio, shao2022decoupled, lablack2023spatio}. Other studies have scaled these models to accommodate larger road networks \cite{mallick2020dynamic, mallick2020graph, zheng2023hybrid}, and integrated advanced techniques like transfer learning \cite{mallick2021transfer, huang2021transfer} and uncertainty quantification \cite{mallick2022deep}. 
Nonetheless, spatiotemporal GNNs struggle to accurately forecast congestion, primarily because the distribution of traffic speed significantly deviates from normality, which compromises the effectiveness of standard loss functions like MAE and MSE. \\
\textbf{Heavy Tail Analysis and Extreme Value Theory.} 
Heavy-tailed random variables, characterized by a high likelihood of significant deviations from the mean, are often observed in road traffic patterns, as exemplified in \cref{fig: 772151}. Statistical techniques in heavy tail analysis often overlap with extreme value theory and have been successful in modelling rare impactful events like floods or heatwaves \cite{haan2006extreme, tanarhte2015heat}. 
In the context of loss functions, \cite{ding2019modeling} introduced Extreme Value Loss (EVL) to predict the future occurrence of extreme events. However, this loss function is based on a binary classification layer, making it difficult to implement for regression problems. To circumvent this, \cite{zhang2021enhancing} proposed a generalized EVL framework inspired by the kernel density estimator (KDE) and introduced the Gumbel Loss and the Frech\'et Loss functions. Nonetheless, Frech\'et Loss is only well-defined only for one-sided extreme events, making it unsuitable for traffic time series data. Meanwhile, \cite{kozerawski2022taming} proposed two moment-based tailedness measurement concepts: Pareto loss and Kurtosis Loss. However, the authors did not specify how to choose the hyperparameters for the generalized Pareto distribution within the Pareto loss; thus, we omit the Pareto loss from our study.
Quantile regression, a method estimating the conditional quantiles of a response variable \cite{koenker2005quantile}, has also been recently adapted into deep learning to model extreme events \cite{wambura2020fast}. These novel loss functions have demonstrated superior efficacy in identifying infrequent events in a time series. It is thus interesting to investigate their potential in improving traffic forecasting. Nonetheless, as remarked in the Introduction, the random variable governing the observed traffic speed, albeit heavy-tailed, does not exactly conform to classical extreme value theory. \\
\textbf{Imbalanced Regression.}
Complimentary to heavy tail analysis, imbalanced regression focuses on restoring a balanced prediction from imbalanced training samples. Although a crucial area, it remains under-explored. 
Recent approaches focus on estimating the prior density distribution of training sets and then reweighting \cite{yang2021delving, steininger2021density}. Yet, it remains uncertain how to effectively estimate the prior density of traffic speed data, given their intricate structure of interdependencies. In the field of computer vision, several loss functions, inspired by Focal Loss, have been proposed and these methods do not require any prior knowledge about the distribution of training labels. Shrinkage Loss utilizes a sigmoid-based function to recalibrate loss terms \cite{lu2018deep}. This approach was later generalized by \cite{yang2021delving}. Most recently, \cite{ren2022balanced} proposed the Balance MSE (bMSE) Loss, which bears a resemblance to the logit adjustment techniques employed in the literature on imbalanced classification. Despite their success in vision-related tasks, the efficacy of these innovative loss functions in traffic forecasting remains an open question, a gap which our study seeks to address.

\section{Methods}

The goal of traffic forecasting is to predict the future traffic speed given previously observed traffic speed from $D$ correlated sensor locations on the road network. Concretely, at each time step, given the representation of a graph $G$ capturing the spatial correlation among the sensor locations and the traffic speed data at $D$ locations for the previous $S$ time steps, the problem is to learn a function $f$ that outputs a 2-dimensional matrix, which represents the traffic speed data at $D$ locations for the next $T$ time steps. For each $d = 1,..., D$ and $t = 1,...,T$, we let $y_{dt}$ denote the observed value at location $d$ and time $t$ and $\hat y_{dt}$ denote the predicted value by the model. In this section, we first define each loss function included in this study. Then, we introduce the metrics that will collectively examine the models' overall performance as well as effectiveness at identifying congestion.

\subsection{Loss Functions}

For our study, we categorize the loss functions into three distinct groups: first-order losses, second-order losses, and Kurtosis Loss.
First-order losses, which impose an L-1 penalty on large errors, include MAE, MAE-Focal, Quantile, and Huber Loss. In general, the L-1 penalty is resistant to values that deviate from the mean.
Second-order losses, applying an L-2 penalty on large errors, consist of MSE, MSE-Focal, Balanced MSE, and Gumbel Loss. Unlike the L-1 penalty, the L-2 penalty adjusts the model to account for ``outliers,'' sometimes at the expense of accuracy on other samples. Except for the balanced MSE Loss, we give the definition of each loss function at the level of each sample.\\
\textbf{Mean absolute error (MAE) and Mean squared error (MSE):} The MAE is defined as:$\frac{1}{D}\frac{1}{T}\sum_{d=1}^D\sum_{t = 1}^T|y_{dt} - \hat y_{dt}|$. The MSE is defined as:$\frac{1}{D}\frac{1}{T}\sum_{d=1}^D\sum_{t = 1}^T(y_{dt} - \hat y_{dt})^2$. In the literature, MAE loss is typically preferred when forecasting traffic with spatiotemporal GNNs because it often induces better overall performance. However, MSE is more sensitive to outliers since it squares the error and exaggerates the effect of outliers.\\
\textbf{Focal Loss:} Focal Loss applies a modulating term to the cross entropy loss in order to focus learning on harder, rarer examples \cite{lin2017focal, lu2018deep}. We adopt a more general variation of the Focal Loss for regression proposed by \cite{yang2021delving}. It is defined as $\frac{1}{D}\frac{1}{T}\sum_{d=1}^D\sum_{t = 1}^T\sigma(|\beta \cdot e_{lt}|)^{\gamma}e_{lt}$,
where $\sigma(\cdot)$ is the sigmoid function, $\beta, \gamma$ are the hyperparameters and $e_{lt}$ is the error at location $l$ and at time $t$. We choose two errors for $e_{lt}$: absolute error: $e_{lt} = |y_{dt} - \hat y_{dt}|$, and squared error: $e_{lt} = (y_{dt} - \hat y_{dt})^2$. We refer to the Focal Loss function with the absolute error as {\bf MAE-Focal} and the one with the squared error as {\bf MSE-Focal}. \\
\textbf{Huber Loss:} Huber Loss is a loss function used in robust regression. It combines MAE with MSE so that it is less sensitive to outliers in data than the MSE \cite{huber1992robust}. Huber Loss is defined as $\sum_{d=1}^D\sum_t^T L_{\beta, d, t}$
    where $$L_{\beta, d, t} = \begin{cases}
    (y_{dt} - \hat y_{dt})^2 /2 & \text{ if } |y_{dt} - \hat y_{dt}| < \beta\\
    |y_{dt} - \hat y_{dt}| - \beta/2 &\text{ otherwise}
\end{cases}$$
and $\beta$ is a hyperparameter.\\
\textbf{Quantile Loss:} Past studies found that Quantile loss can model non-Gaussian and asymmetric patterns in the data \cite{wu2021quantifying,mallick2022deep}. Let $S$ be a set of fixed quantiles between $0$ and $1$. The Quantile Loss is defined as \cite{koenker2005quantile, wu2021quantifying}:$\sum_{\tau \in S} \sum_{d=1}^D\sum_t^T L_{\tau}(y_{dt}, \hat y_{dt})$
    where $L_{\tau}(y_{dt}, \hat y_{dt}) = (y_{dt} - \hat y_{dt}) \cdot (\tau - {\bf 1}(y_{dt} < \hat y_{dt}))$
    for a fixed confidence level $\tau$.\\
\textbf{Balanced MSE:} We adapt the Batch-based Monte-Carlo (BMC) implementation of the Balanced MSE Loss function \cite{ren2022balanced}, which does not impose any assumptions on the label distribution. This loss function takes into account how rare the error is at the batch level. Given a batch of size $B$, we let $\bm y_{bl}$ be a vector of length $T$, which represents the a sequence observed values of $T$ times steps from the $b$th sample in the batch at the $l$th location. The BMC implementation of the Balanced MSE is defined as
$$- \log \frac{\exp(-||\hat{\bm y}_{bl} - \bm y_{bl}||_2^2/(2\sigma^2_{\text{noise}}))}{\sum_{b'=1}^{B}\sum_{l' = 1}^{L}\exp(-||\hat{\bm y}_{bl} - \bm y_{b'l'}||_2^2/(2\sigma^2_{\text{noise}}))}$$
where $\sigma^2_{\text{noise}}$ is a hyperparameter. Even though \cite{ren2022balanced} suggest that $\sigma^2_{\text{noise}}$ can be learnt, it still requires another loss function to choose the best model tested on the validation set; however, this defeats the purpose of our study to find a good loss function both for training and validation. \\
\textbf{Gumbel Loss:} Proposed by \cite{zhang2021enhancing}, the Gumbel Loss is defined as$\frac{1}{T}\sum_{t = 1}^T(1 - \exp(-\delta_{lt}^2))^{\gamma}\delta_{lt}^2$, where $\delta_{lt} = y_{dt} - \hat y_{dt}$ and $\gamma$ is a hyperparameter. Notice that with $\gamma > 1$, it discounts smaller errors more aggressively than the MSE and is thus more sensitive to ``outliers.''\\
\textbf{Kurtosis Loss:} Kurtosis measures the tailedness of a distribution as the scaled fourth moment about the mean. Given a pre-defined loss function $L$, the Kurtosis in \cite{kozerawski2022taming} is defined as $L + \lambda \cdot \left(\frac{\hat L - \mu_{\hat L}}{\sigma_{\hat L}}\right)^4$ where $\hat L$ is the auxiliary loss and $\lambda$ is a hyperparameter. $\mu_{\hat L}$ and $\sigma_{\hat L}$ are the mean and standard deviation of the loss $\hat L$ for a batch of training samples. \cite{kozerawski2022taming} suggest choosing the negative likelihood loss for $\hat L$; however, it is unclear which distribution to choose for the negative likelihood loss. Even though, for example, the generalized Pareto distribution is chosen, it is unclear how to estimate the hyperparameters for traffic speed data. 

\subsection{Evaluation Metrics}

\def\ck{\checkmark}
\begin{table*}[hbt!]
\begin{small}
\centering

\begin{tabular}{l l l  r r r | r r r |r r r}
\toprule
\multirow{2}{*}{Data} & \multirow{2}{*}{Model} & \multirow{2}{*}{Losses}   & \multicolumn{3}{c}{15 min} & \multicolumn{3}{c}{30 min}  & \multicolumn{3}{c}{60 min} \\
\cline{4-6}  \cline{7-9} \cline{10-12}   
&&& {\small MAE} & {\small RMSE} & {\small MAPE} & {\small MAE} & {\small RMSE} & {\small MAPE} & {\small MAE} & {\small RMSE} & {\small MAPE}\\
\midrule
\multirow{20}{*}{\rotatebox[origin=c]{90}{METR-LA}}&\multirow{10}{*}{\rotatebox[origin=c]{90}{GraphWaveNet}}
& MAE & {2.708} & 5.193 & 7.036 & {3.084} & 6.238 & 8.505 & \textbf{3.533} & 7.374 & 10.091 \\
&& MAE-Focal & 2.749 & \textbf{5.120} & 7.111 & 3.127 & \textbf{6.127} & {8.438} & 3.595 & \textbf{7.216} & \textbf{9.982} \\
&& Quantile & \textbf{2.693} & {5.139} & {6.966} & \textbf{3.074} & {6.137} & \textbf{8.393} & {3.552} & {7.278} & 10.032 \\
&& Huber & 2.722 & 5.179 & \textbf{6.931} & 3.129 & 6.273 & 8.451 & 3.580 & 7.392 & {10.026} \\
\cmidrule{3-12}
&& MSE & \textbf{2.870} & {5.117} & {7.438} & \textbf{3.307} & 6.078 & \textbf{8.815} & \textbf{3.846} & 7.069 & \textbf{10.324} \\
&& MSE-Focal & 3.004 & 5.121 & 7.620 & 3.450 & 6.038 & 9.191 & 3.967 & \textbf{7.023} & {10.826} \\
&& bMSE-1 & 2.957 & 5.172 & \textbf{7.294} & 3.412 & 6.091 & {8.832} & 4.060 & 7.232 & 10.889 \\
&& bMSE-9 & 2.982 & 5.153 & 7.673 & 3.400 & {6.031} & 9.350 & 4.193 & 7.142 & 11.730 \\
&& Gumbel & {2.881} & \textbf{5.070} & 7.498 & {3.324} & \textbf{6.017} & 9.069 & {3.846} & {7.028} & 10.870 \\
&& Kurtosis & 2.910 & 5.940 & 7.468 & 3.446 & 7.313 & 9.317 & 4.260 & 9.058 & 12.095 \\
\cmidrule{2-12}
&\multirow{10}{*}{\rotatebox[origin=c]{90}{D2STGNN}}
& MAE & \textbf{2.555} & 4.893 & \textbf{6.485} & \textbf{2.903} & {5.901} & \textbf{7.870} & \textbf{3.350} & 7.080 & 9.757 \\
&& MAE-Focal & 2.595 & \textbf{4.869} & 6.582 & 2.945 & \textbf{5.864} & 7.943 & 3.395 & \textbf{6.956} & \textbf{9.639} \\
&& Quantile & {2.557} & {4.882} & {6.513} & 2.908 & 5.904 & {7.902} & {3.355} & {7.003} & {9.642} \\
&& Huber & 2.559 & 4.904 & 6.541 & {2.906} & 5.914 & 7.944 & 3.360 & 7.114 & 9.936 \\
\cmidrule{3-12}
&& MSE & {2.685} & {4.844} & {6.815} & {3.067} & \textbf{5.769} & {8.211} & {3.633} & {6.787} & {10.101} \\
&& MSE-Focal & 2.927 & 4.955 & 7.358 & 3.318 & 5.911 & 8.803 & 3.775 & 6.911 & 10.357 \\
&& bMSE-1 & 2.705 & 4.895 & 6.879 & 3.095 & 5.820 & 8.261 & 3.657 & 6.842 & 10.350 \\
&& bMSE-9 & 2.823 & 4.968 & 7.073 & 3.175 & 5.844 & 8.389 & 3.890 & 6.906 & 10.639 \\
&& Gumbel & \textbf{2.674} & \textbf{4.837} & \textbf{6.788} & \textbf{3.056} & {5.770} & \textbf{8.201} & \textbf{3.584} & \textbf{6.769} & \textbf{10.070} \\
&& Kurtosis & 2.787 & 5.736 & 6.908 & 3.280 & 7.034 & 8.418 & 4.001 & 8.602 & 10.550 \\
\midrule
\midrule
\multirow{20}{*}{\rotatebox[origin=c]{90}{PEMS-BAY}}&\multirow{10}{*}{\rotatebox[origin=c]{90}{GraphWaveNet}}
& MAE & \textbf{1.310} & 2.748 & \textbf{2.741} & \textbf{1.641} & {3.701} & \textbf{3.674} & \textbf{1.964} & \textbf{4.497} & \textbf{4.594} \\
&& MAE-Focal & 1.336 & \textbf{2.727} & 2.790 & 1.673 & \textbf{3.666} & 3.748 & 2.009 & {4.498} & 4.759 \\
&& Quantile & {1.321} & {2.748} & {2.743} & {1.667} & 3.734 & 3.777 & {1.997} & 4.547 & 4.831 \\
&& Huber & 1.324 & 2.756 & 2.757 & 1.675 & 3.767 & {3.707} & 2.019 & 4.605 & {4.676} \\
\cmidrule{3-12}
&& MSE & {1.374} & {2.714} & {2.925} & \textbf{1.712} & \textbf{3.590} & {3.849} & \textbf{2.035} & \textbf{4.304} & {4.804} \\
&& MSE-Focal & 1.450 & 2.759 & 3.069 & 1.804 & 3.643 & 3.957 & 2.120 & {4.314} & 4.836 \\
&& bMSE-1 & 1.417 & 2.747 & 2.947 & 1.744 & 3.622 & \textbf{3.837} & 2.122 & 4.422 & 4.872 \\
&& bMSE-9 & 1.473 & 2.783 & 3.064 & 1.780 & 3.647 & 3.960 & 2.501 & 4.738 & 5.519 \\
&& Gumbel & \textbf{1.371} & \textbf{2.707} & \textbf{2.846} & {1.724} & {3.617} & 3.863 & {2.044} & 4.369 & \textbf{4.795} \\
&& Kurtosis & 1.870 & 4.892 & 3.899 & 2.238 & 5.543 & 4.873 & 2.708 & 6.370 & 6.012 \\
\cmidrule{2-12}
&\multirow{10}{*}{\rotatebox[origin=c]{90}{D2STGNN}}
& MAE & \textbf{1.253} & 2.631 & \textbf{2.621} & {1.566} & 3.578 & \textbf{3.539} & 1.881 & 4.354 & {4.361} \\
&& MAE-Focal & 1.284 & \textbf{2.623} & 2.663 & 1.604 & \textbf{3.568} & 3.572 & 1.932 & \textbf{4.314} & 4.418 \\
&& Quantile & {1.253} & {2.630} & {2.621} & \textbf{1.565} & 3.573 & {3.545} & \textbf{1.876} & 4.319 & \textbf{4.357} \\
&& Huber & 1.258 & 2.631 & 2.631 & 1.572 & {3.571} & 3.567 & {1.879} & {4.316} & 4.391 \\
\cmidrule{3-12}
&& MSE & {1.348} & {2.688} & {2.851} & {1.666} & \textbf{3.552} & {3.746} & \textbf{1.987} & \textbf{4.259} & \textbf{4.573} \\
&& MSE-Focal & 1.429 & 2.729 & 2.955 & 1.794 & 3.626 & 3.916 & 2.138 & 4.355 & 4.833 \\
&& bMSE-1 & 1.383 & 2.742 & 2.961 & 1.722 & {3.618} & 3.867 & 2.093 & 4.425 & 4.801 \\
&& bMSE-9 & 1.532 & 2.838 & 3.207 & 1.836 & 3.663 & 4.087 & 2.501 & 4.685 & 5.522 \\
&& Gumbel & \textbf{1.322} & \textbf{2.677} & \textbf{2.782} & \textbf{1.659} & 3.622 & \textbf{3.721} & {1.997} & {4.344} & {4.628} \\
&& Kurtosis & 1.392 & 3.103 & 2.970 & 2.092 & 5.162 & 4.476 & 2.893 & 6.820 & 6.114 \\
\bottomrule
\end{tabular}

\caption{This table compares the overall performance of loss functions with MAE, RMSE, MAPE. Against MAE, the MAE-Focal Loss often show similar MAE and MAPE performance but consistently lower RMSE. Meawhile, the Gumbel Loss often outperforms MSE or exhibits similar results.}
\label{table:result_common}

\end{small}
\end{table*}

\begin{table*}[hbt!]
\begin{small}
\centering

\begin{tabular}{l l l  r r r | r r r |r r r}
\toprule
\multirow{2}{*}{Data} & \multirow{2}{*}{Model} & \multirow{2}{*}{Losses}   & \multicolumn{3}{c}{15 min} & \multicolumn{3}{c}{30 min}  & \multicolumn{3}{c}{60 min} \\
\cline{4-6}  \cline{7-9} \cline{10-12}   
&&& {\small MAE} & {\small RMSE} & {\small MAPE} & {\small MAE} & {\small RMSE} & {\small MAPE} & {\small MAE} & {\small RMSE} & {\small MAPE}\\
\midrule
\multirow{20}{*}{\rotatebox[origin=c]{90}{METR-LA}}&\multirow{10}{*}{\rotatebox[origin=c]{90}{GraphWaveNet}}
& MAE & 6.741 & 10.882 & 17.240 & 8.092 & 13.042 & {20.784} & 9.138 & 14.615 & 24.251 \\
&& MAE-Focal & \textbf{6.667} & \textbf{10.610} & \textbf{16.784} & \textbf{7.921} & \textbf{12.606} & \textbf{20.406} & \textbf{8.815} & \textbf{13.932} & \textbf{23.560} \\
&& Quantile & 6.794 & 10.905 & {17.195} & {8.036} & {12.845} & 20.837 & {8.941} & {14.215} & {24.010} \\
&& Huber & {6.730} & {10.754} & 17.620 & 8.058 & 12.880 & 21.095 & 9.032 & 14.346 & 24.477 \\
\cmidrule{3-12}
&& MSE & \textbf{6.765} & \textbf{10.233} & \textbf{16.798} & \textbf{7.904} & 11.909 & 19.777 & 8.725 & 13.100 & 22.851 \\
&& MSE-Focal & 6.985 & 10.486 & 17.323 & 8.120 & 12.157 & 20.019 & 8.765 & {13.030} & 22.019 \\
&& bMSE-1 & 7.049 & 10.557 & 18.848 & 8.105 & 12.053 & 21.205 & 8.811 & 13.061 & 22.753 \\
&& bMSE-9 & 7.026 & 10.437 & 17.668 & 8.053 & {11.898} & \textbf{19.238} & \textbf{8.671} & \textbf{12.702} & \textbf{19.312} \\
&& Gumbel & {6.840} & {10.330} & {16.895} & {7.922} & \textbf{11.883} & {19.288} & {8.712} & 13.058 & {21.167} \\
&& Kurtosis & 8.030 & 13.179 & 22.028 & 9.814 & 15.620 & 27.456 & 10.881 & 16.850 & 30.036 \\
\cmidrule{2-12}
&\multirow{10}{*}{\rotatebox[origin=c]{90}{D2STGNN}}
& MAE & 6.094 & 10.068 & 15.597 & 7.387 & 12.216 & 18.880 & 8.519 & 13.965 & {21.300} \\
&& MAE-Focal & {6.072} & \textbf{9.922} & {15.372} & \textbf{7.337} & \textbf{12.015} & {18.546} & \textbf{8.477} & \textbf{13.704} & 21.985 \\
&& Quantile & \textbf{6.063} & {10.023} & \textbf{15.353} & 7.370 & {12.156} & 18.652 & {8.480} & {13.787} & 21.469 \\
&& Huber & 6.095 & 10.098 & 15.480 & {7.347} & 12.167 & \textbf{18.511} & 8.603 & 14.103 & \textbf{21.160} \\
\cmidrule{3-12}
&& MSE & {6.169} & 9.684 & {15.481} & {7.353} & 11.545 & {18.393} & 8.452 & 12.975 & 21.161 \\
&& MSE-Focal & 6.360 & 9.711 & 15.657 & 7.522 & 11.558 & 18.509 & 8.408 & 12.935 & 21.255 \\
&& bMSE-1 & 6.209 & {9.684} & 15.685 & 7.381 & 11.536 & 18.591 & {8.316} & {12.806} & \textbf{20.106} \\
&& bMSE-9 & 6.326 & 9.739 & 17.182 & 7.409 & {11.504} & 18.672 & 8.378 & \textbf{12.684} & {20.133} \\
&& Gumbel & \textbf{6.146} & \textbf{9.626} & \textbf{15.289} & \textbf{7.322} & \textbf{11.482} & \textbf{18.140} & \textbf{8.308} & 12.809 & 20.524 \\
&& Kurtosis & 7.446 & 12.436 & 20.828 & 9.079 & 14.681 & 26.375 & 10.041 & 15.826 & 29.294 \\
\midrule
\midrule
\multirow{20}{*}{\rotatebox[origin=c]{90}{PEMS-BAY}}&\multirow{10}{*}{\rotatebox[origin=c]{90}{GraphWaveNet}}
& MAE & 3.024 & 5.172 & 6.821 & 3.737 & 6.462 & {8.583} & {4.359} & {7.612} & 10.415 \\
&& MAE-Focal & 3.001 & \textbf{5.071} & {6.692} & \textbf{3.721} & \textbf{6.359} & \textbf{8.361} & 4.404 & 7.696 & {9.928} \\
&& Quantile & {2.982} & {5.072} & 6.875 & 3.756 & {6.445} & 8.729 & \textbf{4.312} & \textbf{7.462} & \textbf{9.783} \\
&& Huber & \textbf{2.939} & 5.093 & \textbf{6.612} & {3.724} & 6.546 & 8.685 & 4.388 & 7.700 & 10.522 \\
\cmidrule{3-12}
&& MSE & {2.994} & 4.913 & \textbf{6.652} & {3.728} & 6.139 & {8.343} & 4.228 & 7.080 & \textbf{9.414} \\
&& MSE-Focal & 3.018 & {4.880} & {6.660} & 3.754 & {6.138} & 8.393 & {4.174} & \textbf{6.880} & 9.455 \\
&& bMSE-1 & 3.051 & 4.942 & 6.979 & 3.742 & 6.175 & 8.578 & 4.254 & 7.171 & 9.714 \\
&& bMSE-9 & 3.166 & 4.982 & 7.205 & 3.783 & 6.165 & 8.461 & 4.653 & 7.368 & 10.374 \\
&& Gumbel & \textbf{2.963} & \textbf{4.869} & 6.683 & \textbf{3.675} & \textbf{6.065} & \textbf{8.268} & \textbf{4.167} & {7.069} & {9.435} \\
&& Kurtosis & 4.262 & 7.966 & 10.870 & 5.299 & 9.417 & 13.089 & 6.267 & 10.677 & 15.287 \\
\cmidrule{2-12}
&\multirow{10}{*}{\rotatebox[origin=c]{90}{D2STGNN}}
& MAE & {2.664} & 4.678 & {5.984} & \textbf{3.320} & {5.910} & {7.596} & {3.853} & 6.796 & {9.008} \\
&& MAE-Focal & 2.679 & \textbf{4.607} & 6.057 & 3.370 & \textbf{5.889} & 7.763 & 3.874 & {6.734} & 9.104 \\
&& Quantile & 2.684 & 4.728 & 6.034 & 3.364 & 6.005 & 7.681 & 3.900 & 6.896 & 9.143 \\
&& Huber & \textbf{2.661} & {4.668} & \textbf{5.974} & {3.337} & 5.925 & \textbf{7.581} & \textbf{3.798} & \textbf{6.662} & \textbf{8.661} \\
\cmidrule{3-12}
&& MSE & {2.928} & 4.820 & {6.513} & {3.592} & 5.980 & {8.043} & \textbf{3.980} & {6.686} & \textbf{9.059} \\
&& MSE-Focal & 2.996 & {4.818} & 6.803 & 3.690 & {5.942} & 8.514 & 4.029 & \textbf{6.518} & 9.374 \\
&& bMSE-1 & 3.081 & 5.002 & 6.980 & 3.702 & 6.071 & 8.490 & 4.105 & 6.791 & 9.419 \\
&& bMSE-9 & 3.206 & 4.988 & 7.099 & 3.667 & \textbf{5.918} & 8.260 & 4.416 & 6.881 & 9.808 \\
&& Gumbel & \textbf{2.784} & \textbf{4.692} & \textbf{6.248} & \textbf{3.518} & 5.989 & \textbf{8.020} & {4.010} & 6.786 & {9.138} \\
&& Kurtosis & 3.459 & 6.270 & 8.145 & 5.089 & 9.160 & 12.630 & 6.252 & 10.758 & 16.103 \\
\bottomrule
\end{tabular}

\caption{This table compares the performance of loss functions with MAE, RMSE, MAPE at identified congestion scenarios. Against MAE, the MAE-Focal Loss often show similar MAE and MAPE performance but consistently lower RMSE. Meawhile, the Gumbel Loss often outperforms MSE or exhibits similar results.} 
\label{table:result_cpi}

\end{small}
\end{table*}

\begin{table*}[hbt!]
\begin{small}
\centering

\begin{tabular}{l l l  r r r | r r r |r r r}
\toprule
\multirow{2}{*}{Data} & \multirow{2}{*}{Model} & \multirow{2}{*}{Losses}   & \multicolumn{3}{c}{15 min} & \multicolumn{3}{c}{30 min}  & \multicolumn{3}{c}{60 min} \\
\cline{4-6}  \cline{7-9} \cline{10-12}   
&&& {\small 95\%} & {\small 98\%} & {\small 99\%} & {\small 95\%} & {\small 98\%} & {\small 99\%} & {\small 95\%} & {\small 98\%} & {\small 99\%}\\
\midrule
\multirow{20}{*}{\rotatebox[origin=c]{90}{METR-LA}}&\multirow{10}{*}{\rotatebox[origin=c]{90}{GraphWaveNet}}
& MAE & 9.769 & 17.009 & 23.599 & 11.843 & 22.101 & 30.149 & 14.971 & 28.111 & 36.373 \\
&& MAE-Focal & \textbf{9.698} & 16.557 & 22.922 & \textbf{11.729} & 21.407 & 29.157 & \textbf{14.658} & 26.873 & 35.021 \\
&& Quantile & \underline{9.722} & 16.789 & 23.221 & \underline{11.759} & 21.467 & 29.154 & \underline{14.771} & 27.169 & 35.437 \\
&& Huber & 9.843 & 16.968 & 23.462 & 12.091 & 22.297 & 30.115 & 15.072 & 27.931 & 36.201 \\
&& MSE & 10.288 & 16.486 & 21.799 & 12.517 & 20.632 & 26.981 & 15.136 & 24.811 & 31.787 \\
&& MSE-Focal & 10.153 & \textbf{16.260} & 21.584 & 12.205 & \textbf{20.069} & 26.574 & 14.910 & \underline{24.300} & \underline{30.965} \\
&& bMSE-1 & 10.422 & 16.689 & 22.034 & 12.769 & 20.407 & \underline{26.396} & 15.772 & 24.886 & 31.438 \\
&& bMSE-9 & 10.354 & 16.436 & \underline{21.571} & 12.606 & \underline{20.093} & \textbf{26.003} & 15.363 & \textbf{24.007} & \textbf{30.399} \\
&& Gumbel & 10.167 & \underline{16.275} & \textbf{21.555} & 12.436 & 20.248 & 26.459 & 15.099 & 24.485 & 31.279 \\
&& Kurtosis & 10.434 & 21.169 & 30.852 & 13.941 & 30.573 & 37.648 & 23.009 & 37.430 & 40.220 \\
\cmidrule{2-12}
&\multirow{10}{*}{\rotatebox[origin=c]{90}{D2STGNN}}
& MAE & 9.103 & 15.586 & 22.027 & \underline{10.773} & 20.420 & 28.683 & \underline{13.636} & 26.805 & 35.645 \\
&& MAE-Focal & \textbf{9.026} & 15.401 & 21.808 & \textbf{10.729} & 20.244 & 28.402 & \textbf{13.452} & 25.951 & 34.728 \\
&& Quantile & \underline{9.083} & 15.566 & 21.934 & 10.819 & 20.415 & 28.736 & 13.676 & 26.167 & 34.729 \\
&& Huber & 9.108 & 15.639 & 22.110 & 10.805 & 20.499 & 28.801 & 13.714 & 26.821 & 35.759 \\
&& MSE & 9.305 & \textbf{15.270} & 20.896 & 11.151 & \textbf{19.357} & \underline{26.477} & 13.993 & 23.871 & 31.327 \\
&& MSE-Focal & 9.547 & 15.430 & \textbf{20.823} & 11.522 & 19.684 & 26.613 & 14.268 & 24.310 & 31.651 \\
&& bMSE-1 & 9.512 & 15.538 & 20.996 & 11.346 & 19.555 & 26.580 & 14.246 & 24.022 & 31.259 \\
&& bMSE-9 & 9.654 & 15.601 & 21.026 & 11.392 & 19.434 & \textbf{26.388} & 14.249 & \textbf{23.727} & \textbf{30.935} \\
&& Gumbel & 9.333 & \underline{15.271} & \underline{20.862} & 11.178 & \underline{19.406} & 26.497 & 13.950 & \underline{23.867} & \underline{31.241} \\
&& Kurtosis & 9.771 & 19.936 & 30.388 & 12.745 & 29.618 & 37.587 & 20.094 & 37.102 & 39.623 \\
\midrule
\midrule
\multirow{20}{*}{\rotatebox[origin=c]{90}{PEMS-BAY}}&\multirow{10}{*}{\rotatebox[origin=c]{90}{GraphWaveNet}}
& MAE & \textbf{4.497} & 8.557 & 12.567 & \textbf{5.956} & 12.124 & 17.971 & \underline{7.403} & 15.426 & 22.478 \\
&& MAE-Focal & 4.545 & \textbf{8.417} & 12.285 & \underline{5.964} & \textbf{11.778} & 17.498 & \textbf{7.402} & 15.139 & 22.307 \\
&& Quantile & 4.537 & 8.560 & 12.522 & 6.070 & 12.293 & 18.134 & 7.500 & 15.533 & 22.584 \\
&& Huber & \underline{4.519} & 8.566 & 12.540 & 5.996 & 12.316 & 18.435 & 7.512 & 15.826 & 23.283 \\
&& MSE & 4.863 & 8.640 & \textbf{12.054} & 6.395 & 11.855 & \underline{16.669} & 7.806 & \underline{14.573} & \underline{20.313} \\
&& MSE-Focal & 5.036 & 8.694 & \underline{12.060} & 6.510 & \underline{11.811} & \textbf{16.587} & 7.839 & \textbf{14.310} & \textbf{20.060} \\
&& bMSE-1 & 4.903 & 8.660 & 12.165 & 6.474 & 11.897 & 16.722 & 8.018 & 14.835 & 20.722 \\
&& bMSE-9 & 4.910 & 8.773 & 12.247 & 6.429 & 12.014 & 17.000 & 8.065 & 15.040 & 21.077 \\
&& Gumbel & 4.743 & \underline{8.490} & 12.066 & 6.339 & 11.833 & 16.832 & 7.749 & 14.720 & 20.785 \\
&& Kurtosis & 6.490 & 23.633 & 29.274 & 9.540 & 26.426 & 29.659 & 14.532 & 28.102 & 30.184 \\
\cmidrule{2-12}
&\multirow{10}{*}{\rotatebox[origin=c]{90}{D2STGNN}}
& MAE & \textbf{4.233} & 8.000 & 11.900 & \textbf{5.536} & 11.244 & 17.149 & \underline{6.864} & 14.421 & 21.641 \\
&& MAE-Focal & 4.306 & \textbf{7.908} & \textbf{11.643} & 5.589 & \textbf{11.112} & 16.849 & 6.910 & \textbf{14.097} & 21.088 \\
&& Quantile & 4.244 & 8.003 & \underline{11.867} & 5.569 & 11.284 & 17.078 & 6.956 & 14.351 & 21.227 \\
&& Huber & \underline{4.239} & \underline{7.969} & 11.892 & \underline{5.553} & \underline{11.237} & 17.037 & \textbf{6.841} & 14.270 & 21.304 \\
&& MSE & 4.710 & 8.464 & 11.986 & 6.135 & 11.493 & \underline{16.417} & 7.482 & 14.204 & \underline{20.142} \\
&& MSE-Focal & 4.908 & 8.495 & 11.931 & 6.492 & 11.647 & \textbf{16.403} & 7.814 & \underline{14.183} & \textbf{20.015} \\
&& bMSE-1 & 4.830 & 8.681 & 12.281 & 6.408 & 11.736 & 16.607 & 7.930 & 14.625 & 20.452 \\
&& bMSE-9 & 5.126 & 8.910 & 12.389 & 6.585 & 11.880 & 16.602 & 8.541 & 14.898 & 20.681 \\
&& Gumbel & 4.546 & 8.266 & 11.890 & 5.994 & 11.536 & 16.893 & 7.482 & 14.497 & 20.754 \\
&& Kurtosis & 4.600 & 9.483 & 15.325 & 8.180 & 23.664 & 29.210 & 17.629 & 29.241 & 30.688 \\
\bottomrule
\end{tabular}

\caption{This table compares the VaR of each loss function at three different levels: 95\%, 98\%, and 99\%. first-order losses yield smaller errors at the $95^{th}$ percentile. However, at the $99^{th}$ percentile, second-order losses are often superior, suggesting they are more effective for managing extreme errors.}
\label{table:result_VaR_AE} 

\end{small}
\end{table*}

\begin{table*}[hbt!]
\begin{small}
\centering

\begin{tabular}{l l | r r | r r r | r r | r r r}
\toprule          
&& \multicolumn{5}{c|}{MAE} & \multicolumn{5}{c}{MSE} \\
\cmidrule{3-12}
&& \multicolumn{2}{c|}{Overall} & \multicolumn{3}{c|}{Congestion} & \multicolumn{2}{c|}{Overall} & \multicolumn{3}{c}{Congestion}\\
\cmidrule{3-12}
Horizon & Loss & MAE & RMSE & MAE & RMSE & \shortstack{Extreme \\VaR Errors} & MAE & RMSE & MAE & RMSE & \shortstack{Extreme\\VaR Errors}\\
\midrule
\multirow{7}{*}{15 min}
& MAE-Focal &\ck   &\ck\ck&\ck\ck&\ck\ck&\ck\ck&\ck\ck&\ck   &\ck\ck&      &  \\
& Quantile  &\ck   &\ck\ck&\ck   &\ck   &\ck\ck&\ck\ck&\ck   &\ck\ck&      &  \\
& Huber     &\ck   &\ck   &\ck\ck&\ck\ck&\ck\ck&\ck\ck&\ck   &\ck\ck&      &  \\
& MSE-Focal &      &\ck\ck&      &      &\ck\ck&      &      &      &\ck   &\ck\ck\\
& bMSE      &      &      &      &      &      &\ck   &\ck   &      &      &      \\
& Gumbel    &      &\ck\ck&      &\ck\ck&\ck\ck&\ck\ck&\ck\ck&\ck\ck&\ck\ck&\ck\ck\\
& Kurtosis  &      &      &      &      &      &      &      &      &      &  \\
\midrule
\multirow{7}{*}{30 min}
& MAE-Focal &\ck   &\ck\ck&\ck\ck&\ck\ck&\ck\ck&\ck\ck&\ck   &\ck   &      &  \\
& Quantile  &\ck   &\ck   &\ck   &\ck   &      &\ck\ck&      &      &      &  \\
& Huber     &\ck   &\ck   &\ck   &\ck\ck&      &\ck\ck&      &      &      &  \\
& MSE-Focal &      &\ck   &      &\ck   &\ck\ck&      &      &      &      &\ck\ck\\
& bMSE      &      &\ck\ck&      &\ck   &\ck\ck&\ck   &\ck   &      &\ck\ck&\ck\\
& Gumbel    &      &\ck\ck&      &\ck\ck&\ck\ck&\ck   &\ck   &\ck\ck&\ck\ck&\ck\\
& Kurtosis  &      &      &      &      &      &      &      &      &      &  \\
\midrule
\multirow{7}{*}{1 hour}
& MAE-Focal &\ck   &\ck\ck&\ck   &\ck\ck&      &\ck\ck&      &      &      &  \\
& Quantile  &\ck   &\ck\ck&\ck\ck&\ck\ck&      &\ck\ck&      &      &      &  \\
& Huber     &\ck   &\ck   &\ck   &      &      &\ck\ck&      &      &      &  \\
& MSE-Focal &      &\ck\ck&      &\ck\ck&      &      &      &\ck   &\ck\ck&\ck\ck\\
& bMSE      &      &\ck\ck&      &\ck\ck&      &      &      &\ck   &\ck   &\ck  \\
& Gumbel    &      &\ck\ck&      &\ck\ck&      &\ck   &\ck   &\ck\ck&\ck\ck&\ck  \\
& Kurtosis  &      &      &      &      &      &      &      &      &      &  \\
\bottomrule
\end{tabular}

\caption{Comparison of different loss functions benchmarked against MAE and MSE Losses. A single check mark denotes error rates comparable to or better than MAE or MSE, with a difference of up to $0.1$. Double check marks signify consistently lower errors, except for at most 1 entry. The $99^{th}$ quantile of VaR errors is considered extreme. {\bf Against MAE, the MAE-Focal Loss shows similar MAE performance but consistently lower RMSE. Against MSE, the Gumbel Loss has similar MAE and RMSE results but lower errors during congestion periods.}}
\label{tb:comparison-table}

\end{small}
\end{table*}

We evaluate all models for 15 minutes (3 steps), 30 minutes (6 steps) and 1 hour (12 steps) ahead forecasting. 
We first consider three traditional evaluation metrics for traffic forecasting: Mean Absolute Error (MAE)
, Root Mean Squared Error (RMSE)
, and Mean Absolute Percentage Error (MAPE) \cite{li2018diffusion}. 
These metrics enable us to gauge the average predictive performance over the entire evaluation timeframe. In addition to these standard performance metrics, we introduce metrics designed to capture the model's performance during abrupt speed fluctuations due to traffic congestion. \\
\textbf{Identifying Bimodality in a Time Series.} The first step to capture congestion is to focus on sensor locations that consistently display congestion patterns. To achieve this, we assess the bimodality in the empirical distribution of traffic speed time series data from each location. First, we utilize Kernel Density Estimation (KDE) to smooth the histogram of historical traffic speed. After smoothing, we locate all the local minima that are distanced at least 10 units (mph) from the mode of this curve. For each identified local minima, we compute the percentage of historical speeds below the speed represented by the minima. A time series is deemed to exhibit a {\it significant} bimodal distribution if a local minima exists and the calculated proportion exceeds 0.1. \\
\textbf{Errors in Congested Scenarios:} After all the sensor locations with significant bimodal distribution patterns are identified, we employ offline change point detection to identify the time steps marking the changes in traffic speed \cite{aminikhanghahi2017survey}. 
In particular, 
we employ the linearly penalized segmentation algorithm \cite{killick2012optimal} with the radial basis function (RBF) kernel \cite{garreau2018consistent, arlot2019kernel} provided by the \texttt{ruptures} Python package \cite{truong2020selective}.
Traffic speed changes can span across an interval, while change point detection methods may only identify a point within such an interval. To mitigate this issue, we introduce the ``change point intervals,'' where we incorporate two-time steps preceding and following each identified change point. Upon the identification of change point intervals, we calculate the MAE, RMSE, and MAPE to assess performance at these intervals.\\
\textbf{Value-at-Risk Metric:}
We adapt the Value-at-Risk (VaR) \cite{resnick2007heavy, kozerawski2022taming} metric commonly used in heavy-tail analysis: given $\alpha \in (0, 1)$, the VaR at the level $\alpha$ is defined as:
$\text{VaR}_{\alpha}(E) = \inf\{e \in E: \mathbb{P}(E \ge e) \le 1 - \alpha\}$
i.e. the smallest error $e$ such that the probability of observing error larger than $e$ is smaller than $1-\alpha$, where $E$ is the empirical distribution of error in the test set. This reports the $\alpha$-th quantile of the error distribution. We use the absolute error: $|y_{cdt} - \hat y_{cdt}|$ for $e$ and measure VaR at three different levels: 95\%, 98\%, and 99\%. 

\section{Experiments}

\textbf{Datasets:}
Our research is based on two traffic speed datasets, METR-LA \cite{METR-LA} and PEMS-BAY \cite{PEMS-BAY}, first benchmarked by \cite{li2018diffusion}. METR-LA contains observations from $D = 207$ sensors over 4 months and 82 of the 207 (39.6\%) sensors exhibit a significant bimodal distribution in their historical speeds. PEMS-BAY contains observations from $D = 325$ sensors in the Bay Area over 6 months and 79 of the 325 (24.3\%) sensors exhibit a significant bimodal distribution in their historical speeds. Overall, the traffic presented in PEMS-BAY is less congested than that in METR-LA.\\
\textbf{Experiment Setup:}
We choose two state-of-the-art models for traffic speed forecasting for this study: GraphWaveNet \cite{wu2019graph} and D2STGNN \cite{shao2022decoupled}. We adhere to the same setup as described in their original papers. 
We remark that GraphWaveNet is executable with Pytorch version between 1.3.1 and 1.9.0 and is trained for 100 epochs. D2STGNN is executable with Pytorch version 1.9.1 or later and is trained for 80 epochs. All experiments are carried out on an NVIDIA DGX A100.\\
\textbf{Implementation Details:}
Based on the original implementation, we select the following specifications for  loss functions: Focal Loss $\beta=0.2$ and $\gamma=1$; Quantile Loss \cite{wu2021quantifying} $S = \{0.025, 0.5, 0.975\}$; Huber Loss $\beta = 1$; Balanced MSE: $\sigma^2_{\text{noise}}$ to be $1$ and $9$  (we refer the Balanced MSE Loss with $\sigma^2_{\text{noise}} = 1$ as bMSE-1 and the other as bMSE-9); Gumbel Loss: $\gamma = 1.1$; Kurtosis Loss:  $l$ is the MAE loss, $\hat l$ is the MSE Loss and $\lambda = 0.01$.
\subsection{Results}

\cref{table:result_common} presents a comparison of the overall performance of various loss functions. We divide the first-order losses from second-order losses and align Kurtosis Loss with the second-order loss for a more straightforward visualization. The best-performing loss function within each group is highlighted in bold. We observe that first-order losses generally perform better in MAE, while second-order losses excel in RMSE. Moreover, the MAE-Focal  Loss often outperforms the standard MAE Loss in terms of RMSE. The difference in MAE values between the two can be as significant as 0.1. Meanwhile, Gumbel Loss often displays superior performance to Mean Square Error (MSE) in terms of RMSE, and sometimes in terms of MAE.

\cref{table:result_cpi} is similar to \cref{table:result_common}, except that the prior evaluates the performance of these loss functions specifically in identified traffic congestion scenarios. We observe a notably higher performance of MAE-Focal Loss compared to standard MAE Loss. This can be explained by MAE-Focal Loss placing more emphasis on penalizing rare labels, representative of traffic speeds during congested periods. Meanwhile, the Gumbel Loss tends to exhibit a lower RMSE compared to MSE more frequently. This is due to its factor, $(1 - \exp(-\delta_{lt}^2))^{\gamma}$, which significantly discounts smaller errors, pushing the model to focus on larger errors. Moreover, for both loss functions, improvements on the PEMS-BAY dataset are less pronounced, possibly because traffic in the Bay area is generally less congested.

Interestingly, in congested scenarios (\cref{table:result_cpi}), second-order losses can occasionally outperform in MAE for longer durations (30 min and 1 hour). This can be attributed to that as predictions extend further into the future, second-order losses impose greater penalties on large errors, leading to more aggressive corrections.

\cref{table:result_VaR_AE} compares the Value-at-Risk (VaR) of each loss function across three levels: 95\%, 98\%, and 99\%. We do not group the loss functions for easier visualization. The bold font indicates the best performer and the underline indicates the second best. First-order losses generally yield smaller errors at the $95^{th}$ percentile, indicating better performance for typical speed  observations. Also, the MAE-Focal Loss often outperforms the standard MAE Loss at the $98^{th}$ and $99^{th}$ percentiles. On the other hand, second-order losses tend to have smaller errors at the $99^{th}$ percentile, hinting at their potential in mitigating larger errors. Among them, MSE-Focal tend to produce lower errors across various time horizons. However, no second-order loss consistently outperforms others, making the search for a loss function resilient to large errors in traffic speed data an interesting research question.

As a summary, we benchmark all loss functions against MAE and MSE in \cref{tb:comparison-table}: {\bf against MAE, MAE-Focal Loss is recommended due to its comparable MAE but consistently lower RMSE in both general and congested scenarios; against MSE, the Gumbel Loss is recommended, given its similar MAE and RMSE in general scenarios but superior performance during congested periods.}

\section{Conclusion}
In this study, we perform benchmark analyses to assess the efficacy of multiple loss functions in traffic forecasting, emphasizing their ability to forecast congestion, a significant challenge faced by existing AI systems. These evaluations are carried out on two datasets, META-LA and PEMS-BAY, leading us to the following recommendation: for objectives centered on the optimization of MAE, we recommend the MAE-Focal Loss function; for objectives directed toward the optimization of MSE, we recommend the Gumbel Loss. These loss functions enhance deep learning models' efficacy in predicting traffic congestion by incorporating techniques from imbalanced regression and extreme value theory. 

For future work, a crucial area for improvement lies in optimizing the hyperparameter selection, as a more refined hyperparameter tuning approach can help fully harness the capabilities of these novel loss functions. Moreover, heavy tail analysis in the context of traffic speed forecasting is under explored. Thus, one interesting research question is adapting the Generalised Pareto distribution to account for the complex spatiotemporal dependencies in the traffic speed data.

It's imperative to underline that by enhancing the prediction accuracy for congestion, we're paving the way for AI systems that are not only more accurate but also safe, robust, and responsible in real-world applications. 



\section*{Acknowledgments}
This material is based on research supported by a project under the Sustainable Research Pathways (SRP) Program. This research used resources from the Argonne Leadership Computing Facility, which is a DOE Office of Science User Facility under contract DE-AC02-06CH11357. The authors would like to extend their gratitude to Prasanna Balaprakash and Ngoc Mai Tran for their valuable suggestions regarding the methodology of this study.

\section*{Government license}
The submitted manuscript has been created by UChicago Argonne, LLC, Operator of Argonne National Laboratory ("Argonne"). Argonne, a U.S. Department of Energy Office of Science laboratory, is operated under Contract No. DE-AC02-06CH11357. The U.S. Government retains for itself, and others acting on its behalf, a paid-up nonexclusive, irrevocable worldwide license in said article to reproduce, prepare derivative works, distribute copies to the public, and perform publicly and display publicly, by or on behalf of the Government.  The Department of Energy will provide public access to these results of federally sponsored research in accordance with the DOE Public Access Plan. \url{http://energy.gov/downloads/doe-public-access-plan}.

\bibliographystyle{unsrtnat}
\bibliography{references}
\end{document}